\def\year{2018}\relax
\documentclass[letterpaper]{article} 
\usepackage{aaai18}  
\usepackage{times}  
\usepackage{helvet}  
\usepackage{courier}  
\usepackage{url}  
\usepackage{graphicx}  
\usepackage[boxed]{algorithm2e}
\usepackage{amsmath,amssymb}
\usepackage{ctable}
\usepackage{hhline}
\usepackage{xcolor}
\usepackage{colortbl}
\usepackage{multirow}
\usepackage{subfigure}
\graphicspath{{./}{Figures/}}
\def\x{{\mathbf x}}
\definecolor{betterblue}{HTML}{2F5597}
\definecolor{betterorange}{HTML}{C55A11}

\begin{document}
%
\title{Attend and Diagnose: Clinical Time Series Analysis using Attention Models}
\author{ Huan Song$^{\dagger}$, Deepta Rajan$^{\ddagger}$ \thanks{The first two authors contributed equally.}, Jayaraman J. Thiagarajan$^{\dagger\dagger}$, Andreas Spanias$^{\dagger}$ \\
$^{\dagger}$ SenSIP Center, School of ECEE, Arizona State University, Tempe, AZ \\
$^{\ddagger}$ IBM Almaden Research Center, 650 Harry Road, San Jose, CA \\
$^{\dagger\dagger}$ Lawrence Livermore National Labs, 7000 East Avenue, Livermore, CA
}
\maketitle
\begin{abstract}
With widespread adoption of electronic health records, there is an increased emphasis for predictive models that can effectively deal with clinical time-series data. Powered by Recurrent Neural Network (RNN) architectures with Long Short-Term Memory (LSTM) units, deep neural networks have achieved state-of-the-art results in several clinical prediction tasks. Despite the success of RNNs, its sequential nature prohibits parallelized computing, thus making it inefficient particularly when processing long sequences. Recently, architectures which are based solely on attention mechanisms have shown remarkable success in transduction tasks in NLP, while being computationally superior. In this paper, for the first time, we utilize attention models for clinical time-series modeling, thereby dispensing recurrence entirely. We develop the \textit{SAnD} (Simply Attend and Diagnose) architecture, which employs a masked, self-attention mechanism, and uses positional encoding and dense interpolation strategies for incorporating temporal order. Furthermore, we develop a multi-task variant of \textit{SAnD} to jointly infer models with multiple diagnosis tasks. Using the recent MIMIC-III benchmark datasets, we  demonstrate that the proposed approach achieves state-of-the-art performance in all tasks, outperforming LSTM models and classical baselines with hand-engineered features.
\end{abstract}

\section{Introduction}
Healthcare is one of the prominent applications of data mining and machine learning, and it has witnessed tremendous growth in research interest recently. This can be directly attributed to both the abundance of digital clinical data, primarily due to the widespread adoption of electronic health records (EHR), and advances in data-driven inferencing methodologies. Clinical data, for example intensive care unit (ICU) measurements, is often comprised of multi-variate, time-series observations corresponding to sensor measurements, test results and subjective assessments. Potential inferencing tasks using such data include classifying diagnoses accurately, estimating length of stay, and predicting future illness, or mortality. 

The classical approach for healthcare data analysis has been centered around extracting hand-engineered features and building task-specific predictive models. Machine learning models are often challenged by factors such as need for long-term dependencies, irregular sampling and missing values.  In the recent years, recurrent Neural Networks (RNNs) based on Long Short-Term Memory (LSTM) \cite{hochreiter1997long} have become the de facto solution to deal with clinical time-series data. RNNs are designed to model varying-length data and have achieved state-of-the-art results in sequence-to-sequence modeling \cite{sutskever2014sequence}, image captioning \cite{xu2015show} and recently in clinical diagnosis \cite{lipton2015learning}. Furthermore, LSTMs are effective in exploiting long-range dependencies and handling nonlinear dynamics. 


\subsubsection{Attention in Clinical Data Analysis:}RNNs perform computations at each position of the time-series by generating a sequence of hidden states as a function of the previous hidden state and the input for current position. This inherent sequential nature makes parallelization challenging. Though efforts to improve the computational efficiency of sequential modeling have recently surfaced, some of the limitations still persist. The recent work of Vaswani \textit{et. al.} argues that attention mechanisms, without any recurrence, can be effective in sequence-to-sequence modeling tasks. Attention mechanisms are used to model dependencies in sequences without regard for their actual distances in the sequence \cite{bahdanau2014neural}. 

In this paper, we develop \textit{SAnD} (Simply Attend and Diagnose), a new approach for clinical time-series analysis, which is solely based on attention mechanisms. In contrast to sequence-to-sequence modeling in NLP, we propose to use \textit{self-attention} that models dependencies within a single sequence. In particular, we adopt the \textit{multi-head} attention mechanism similar to \cite{vaswani2017attention}, with an additional masking to enable causality. In order to incorporate temporal order into the representation learning, we propose to utilize both positional encoding and a dense interpolation embedding technique.

\subsubsection{Evaluation on MIMIC-III Benchmark Dataset:}Another important factor that has challenged machine learning research towards clinical diagnosis is the lack of universally accepted benchmarks to rigorously evaluate the modeling techniques. Consequently, in an effort to standardize research in this field, in \cite{harutyunyan2017multitask}, the authors proposed public benchmarks for four different clinical tasks: mortality prediction, detection of physiologic decompensation, forecasting length of stay, and phenotyping. Interestingly, these benchmarks are supported by the Medical Information Mart for Intensive Care (MIMIC-III) database \cite{johnson2016mimic}, the largest publicly available repository of rich clinical data currently available. These datasets exhibit characteristics that are typical of any large-scale clinical data, including varying-length sequences, skewed distributions and missing values. In \cite{lipton2015learning,harutyunyan2017multitask}, the authors established that RNNs with LSTM cells outperformed all existing baselines including methods with engineered features. 

In this paper, we evaluate \textit{SAnD} on all MIMIC-III benchmark tasks and show that it is highly competitive, and in most cases outperforms the state-of-the-art LSTM based RNNs. Both superior performance and computational efficiency clearly demonstrate the importance of attention mechanisms in clinical data.

\subsubsection{Contributions:}Here is a summary of our contributions:
\begin{itemize}
	\item We develop the first attention-model based architecture for processing multi-variate clinical time-series data.
	\item Based on the multi-head attention mechanism in \cite{vaswani2017attention}, we design a masked self-attention modeling unit for sequential data.
	\item We propose to include temporal order into the sequence representation using both positional encoding and a dense interpolation technique.
	\item We rigorously evaluate our approach on all MIMIC-III benchmark tasks and achieve state-of-the-art prediction performance.
	\item Using a multi-task learning study, we demonstrate the effectiveness of the \textit{SAnD} architecture over RNNs in joint inferencing.
\end{itemize}



\section{Related Work}
Clinical data modeling is inherently challenging due to a number of factors : a) irregular sampling. b) missing values and measurement errors. c) heterogeneous measurements obtained at often misaligned time steps and presence of long-range dependencies. A large body of work currently exists designed to tackle these challenges -- the most commonly utilized ideas being Linear Dynamical System (LDS) and Gaussian Process (GP). As a classic tool in time-series analysis, LDS models the linear transition between consecutive states \cite{liu2013clinical,liu2016learning}. LDS can be augmented by GP to provide more general non-linear modeling on local sequences, thereby dealing with the irregular sampling issue \cite{liu2013clinical}. In order to handle the multi-variate nature of measurements, \cite{ghassemi2015multivariate} proposed a multi-talk GP method which jointly transforms the measurements into a unified latent space.

More recently, RNNs have become the sought-after solution for clinical sequence modeling. The earliest effort was by Lipton \textit{et. al.} \cite{lipton2015learning}, which propose to use LSTMs with additional training strategies for diagnosis tasks. In \cite{lipton2016modeling}, RNNs are demonstrated to automatically deal with missing values when they are simply marked by an indicator. In order to learn representations that preserve spatial, spectral and temporal patterns, recurrent convolutional networks have been used to model EEG data in \cite{bashivan2015learning}. After the introduction of the MIMIC-III datasets, \cite{harutyunyan2017multitask} have rigorously benchmarked RNNs on all four clinical prediction tasks and further improved the RNN modeling through joint training on all tasks. 

Among many RNN realizations in NLP, attention mechanism is an integral part, often placed between LSTM encoder and decoder \cite{bahdanau2014neural,xu2015show,vinyals2015grammar,hermann2015teaching}.  Recent research in language sequence generation indicates that by stacking the blocks of solely attention computations, one can achieve similar performance as RNN \cite{vaswani2017attention}. In this paper, we propose the first attention based sequence modeling architecture for multivariate time-series data, and study their effectiveness in clinical diagnosis.

\section{Proposed Approach}
\begin{figure*}[t]
	\centering
	\centerline{\includegraphics[width=0.99\linewidth]{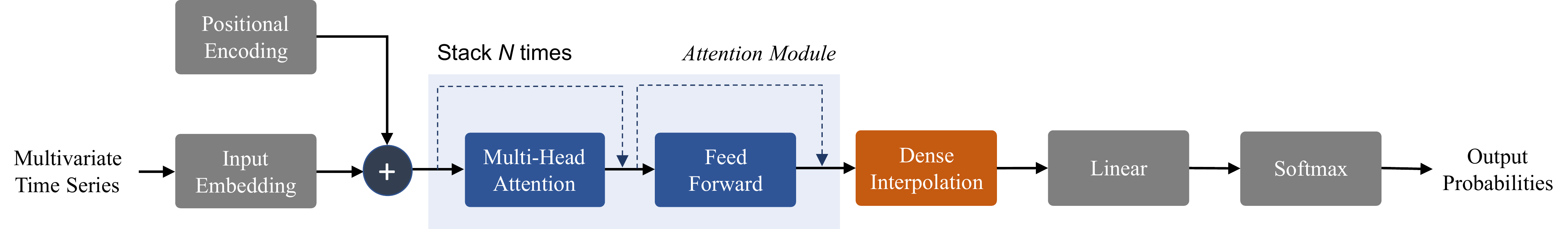}}
	\caption{An overview of the proposed approach for clinical time-series analysis. In contrast to state-of-the-art approaches, this does not utilize any recurrence or convolutions for sequence modeling. Instead, it employs a simple self-attention mechanism coupled with a dense interpolation strategy to enable sequence modeling. The attention module is comprised of $N$ identical layers, which in turn contain the attention mechanism and a feed-forward sub-layer, along with residue connections.}
	\label{fig:architecture}
\end{figure*}

In this section, we describe \textit{SAnD}, a fully attention mechanism based approach for multivariate time-series modeling. The effectiveness of LSTMs have been established in a wide-range of clinical prediction tasks. In this paper, we are interested in studying the efficacy of attention models in similar problems, dispensing recurrence entirely. While core components from the Transformer model \cite{vaswani2017attention} can be adopted, key architectural modifications are needed to solve multivariate time-series inference problems.

The motivation for using attention models in clinical modeling is three-fold: (i) \textit{Memory}: While LSTMs are effective in sequence modeling, lengths of clinical sequences are often very long and in many cases they rely solely on short-term memory to make predictions. Attention mechanisms will enable us to understand the amount of memory modeling needed in benchmark tasks for medical data; (ii) \textit{Optimization}: The mathematical simplicity of attention models will enable the use of additional constraints, e.g. explicit modeling of correlations between different measurements in data, through inter-attention; (iii) \textit{Computation}: Parallelization of sequence model training is challenging, while attention models are fully parallelizable.

\subsection{Architecture}
Our architecture is inspired by the recent Transformer model for sequence transduction \cite{vaswani2017attention}, where the encoder and decoder modules were comprised solely of an attention mechanism. The Transformer architecture achieves superior performance on machine translation benchmarks, while being significantly faster in training when compared to LSTM-based recurrent networks \cite{sutskever2014sequence,wu2016google}. Given a sequence of symbol representations (e.g.\ words) ($x_1, \dots, x_T$), the encoder transforms them into a continuous representation $\mathbf{z}$ and then the decoder  produces the output sequence ($y_1, \dots, y_T$) of symbols.

Given a sequence of clinical measurements ($\x_1, \dots, \x_T$), $\x_t\in \mathbb{R}^{R}$ where $R$ denotes the number of variables, our objective is to generate a sequence-level prediction. The type of prediction depends on the specific task and can be denoted as a discrete scalar $y$ for multi-class classification, a discrete vector $\mathbf{y}$ for multi-label classification and a continuous value $y$ for regression problems. The proposed architecture is shown in Figure \ref{fig:architecture}. In the rest of this section, we describe each of the components in detail.

\subsubsection{Input Embedding:}Given the $R$ measurements at every time step $t$, the first step in our architecture is to generate an embedding that captures the dependencies across different variables without considering the temporal information. This is conceptually similar to the input embedding step in most NLP architectures, where the words in a sentence are mapped into a high-dimensional vector space to facilitate the actual sequence modeling \cite{kim2014convolutional}. To this end, we employ a $1$D convolutional layer to obtain the $d$-dimensional ($d>R$) embeddings for each $t$. Denoting the convolution filter coefficients as $\mathbf{w}\in \mathbb{R}^{T\times h}$, where $h$ is the kernel size, we obtain the input embedding: $\mathbf{w}\cdot \x_{i:i+h-1}$ for the measurement position $i$.

\subsubsection{Positional Encoding:}Since our architecture contains no recurrence, in order to incorporate information about the order of the sequence, we include information about the relative or absolute position of the time-steps in the sequence. In particular, we add \textit{positional encodings} to the input embeddings of the sequence. The encoding is performed by mapping time step $t$ to the same randomized lookup table during both training and prediction. The $d$-dimensional positional embedding is then added to the input embedding with the same dimension. Note that, there are alternative approaches to positional encoding, including the sinusoidal functions in \cite{vaswani2017attention}. However, the proposed strategy is highly effective in all our tasks.

\subsubsection{Attention Module:}Unlike transduction tasks in NLP, our inferencing tasks often require classification or regression architectures. Consequently, \textit{SAnD} relies almost entirely on self-attention mechanisms. Self-attention, also referred as intra-attention, is designed to capture dependencies of a single sequence. Self-attention has been used successfully in a variety of NLP tasks including reading comprehension \cite{cui2016attention} and abstractive summarization \cite{paulus2017deep}. As we will describe later, we utilize a restricted self-attention that imposes causality, i.e., considers information only from positions earlier than the current position being analyzed. In addition, depending on the task we also determine the range of dependency to consider. For example, we will show in our experiments that phenotyping tasks require a longer range dependency compared to mortality prediction.

In general, an attention function can be defined as mapping a query $\mathbf{q}$ and a set of key-value pairs $\{\mathbf{k}, \mathbf{v}\}$ to an output $\mathbf{o}$. For each position $t$, we compute the attention weighting as the inner product between $\mathbf{q}_t$ and keys at every other position in the sequence (within the restricted set) $\{\mathbf{k}_{t'}\}_{t'=t-r}^{t-1}$, where $r$ is the mask size. Using these attention weights, we compute $\mathbf{o}$ as weighted combination of the value vectors $\{\mathbf{v}_{t'}\}_{t'=t-r}^{t-1}$ and pass $\mathbf{o}$ through a feed-forward network to obtain the vector representation for $t$. Mathematically, the attention computation can be expressed as follows:
\begin{equation}
\text{Attention}(\mathbf{Q}, \mathbf{K}, \mathbf{V})=\text{softmax}\left(\frac{\mathbf{Q}\mathbf{K}^\mathbf{T}}{\sqrt{d}}\right)\mathbf{V},
\label{eq:attention}
\end{equation}where $\mathbf{Q}, \mathbf{K}, \mathbf{V}$ are the matrices formed by query, key and value vectors respectively, and $d$ is the dimension of the key vectors. This mechanism is often referred to as the scalar dot-product attention. Since we use only self-attention, $\mathbf{Q}, \mathbf{K}, \mathbf{V}$ all correspond to input embeddings of the sequence (with position encoding). Additionally, we mask the sequence to specify how far the attention models can look into the past for obtaining the representation for each position. Hence, to be precise, we refer to this as \textit{masked self-attention}.

Implicitly, self-attention creates a graph structure for the sequence, where edges indicate the temporal dependencies. Instead of computing a single attention graph, we can actually create multiple attention graphs each of which is defined by different parameters. Each of these attention graphs can be interpreted to encode different types of edges and hence can provide complementary information about different types of dependencies. Hence, we use ``multi-head attention'' similar to \cite{vaswani2017attention}, where $8$ heads are used to create multiple attention graphs and the resulting weighted representations are concatenated and linearly projected to obtain the final representation. The second component in the attention module is $1$D convolutional sub-layers with kernel size $1$, similar to the input embedding. Internally, we use two of these $1$D convolutional sub-layers with ReLU (rectified linear unit) activation in between. Note that, we include residue connections in both the sub-layers. 

Since we stack the attention module $N$ times, we perform the actual prediction task using representations obtained at the final attention module. Unlike transduction tasks, we do not make predictions at each time step in all cases. Hence, there is a need to create a concise representation for the entire sequence using the learned representations, for which we employ a dense interpolated embedding scheme, that encodes partial temporal ordering. 

\subsubsection{Dense Interpolation for Encoding Order:}The simplest approach to obtain a unified representation for a sequence, while preserving order, is to simply concatenate embeddings at every time step. However, in our case, this can lead to a very high-dimensional, ``cursed'' representation which is not suitable for learning and inference. Consequently, we propose to utilize a  dense interpolation algorithm from language modeling. Besides providing a concise representation, \cite{trask2015modeling} demonstrated that the dense interpolated embeddings better encode word structures which are useful in detecting syntactic features. In our architecture, dense interpolation embeddings, along with the positional encoding module, are highly effective in capturing enough temporal structure required for even challenging clinical prediction tasks.

\begin{algorithm}[h]

 \SetKwInOut{Input}{Input}
 \SetKwInOut{Output}{Output}
 \underline{Dense Interpolation Embedding}

 \Input{Steps $t$ of the time series and length of the sequence $T$, embeddings at step $t$ as $\mathbf{s}_t$, factor $M$.}
 \Output{Dense interpolated vector representation $\mathbf{u}$.}
 \For{$t=1$ to $T$}{
 	$s=M*t/T$ \\
 	\For{$m=1$ to $M$}{
 		$w=pow(1-abs(s-m)/M, 2)$
 		$\mathbf{u}_m=\mathbf{u}_m+w*\mathbf{s}_t$
 	}
 }
 \caption{Dense interpolation embedding with partial order for a given sequence.}
 \label{alg:dense_interp}
\end{algorithm}
 \vspace{-10pt}
 
\begin{figure}[t]
	\centering
	\centerline{\includegraphics[width=0.7\linewidth]{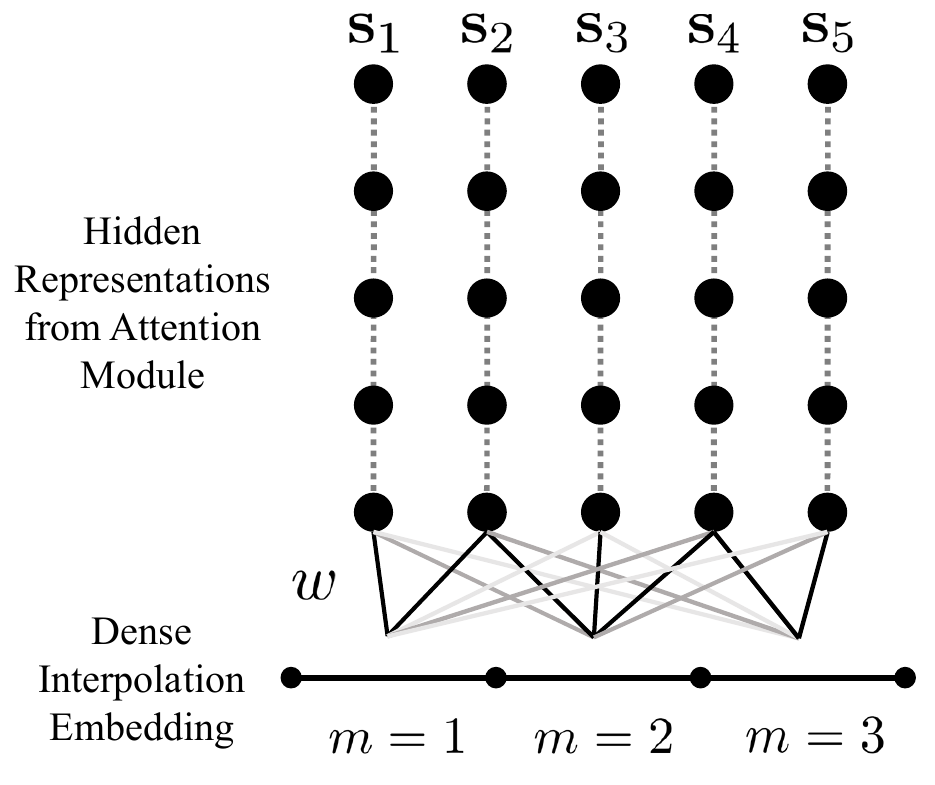}}
	\caption{Visualizing the dense interpolation module, for the case when $T = 5$ and $M = 3$.}
	\label{fig:dense_interp}
	\vspace{-10pt}
\end{figure}

The pseudocode to perform dense interpolation for a given sequence is shown in Algorithm \ref{alg:dense_interp}. Denoting the hidden representation at time $t$, from the attention model, as $\mathbf{s}_t\in \mathbb{R}^{d}$, the interpolated embedding vector will have dimension $d\times M$, where $M$ is the \textit{dense interpolation factor}. Note that when $M=T$, it reduces to the concatenation case. The main idea of this scheme is to determine weights $w$, denoting the contribution of $\mathbf{s}_t$ to the position $m$ of the final vector representation $\mathbf{u}$. As we iterate through the time-steps of a sequence, we obtain $s$, the relative position of time step $t$ in the final representation $\mathbf{u}$ and $w$ is computed as $w=(1-\frac{|s-m|}{M})^2$. We visualize the dense interpolation process in Figure \ref{fig:dense_interp} for the toy case of $T=5, M=3$. The larger weights in $w$ are indicated by darker edges while the lighter edges indicates lesser influence. In practice, dense interpolation is implemented efficiently by caching $w$'s into a matrix $\mathbf{W}\in \mathbb{R}^{T\times M}$ and then performing the following matrix multiplication: $\mathbf{U}=\mathbf{S}\times \mathbf{W}$, where $\mathbf{S}=[\mathbf{s}_1,\dots,\mathbf{s}_T]$. Finally we can obtain $\mathbf{u}$ by stacking columns of $\mathbf{U}$.

\subsubsection{Linear and Softmax layers:}After obtaining a single vector representation from dense interpolation, we utilize a linear layer to obtain the logits. The final layer depends on the specific task. We can use a softmax layer for the binary classification problems, a sigmoid layer for multi-label classification since the classes are not mutually exclusive and a ReLU layer for regression problems. The corresponding loss functions are:

\begin{itemize}
	\item \textbf{Binary classification}: $-(y\cdot \log(\hat{y}))+(1-y)\cdot \log(1-\hat{y})$, where $y$ and $\hat{y}$ are the true and predicted labels.
	\item \textbf{Multi-label classification}: $\frac{1}{K}\sum_{k=1}^K -(y_k\cdot \log(\hat{y}_k)+(1-y_k)\cdot \log(1-\hat{y}_k))$, where $K$ denotes the total number of labels in the dataset.
	\item \textbf{Regression}: $\sum_{t=1}^T (l_t-\hat{l_t})^2$, where $l_t$ and $\hat{l}_t$ denote the true and predicted response variables at time-step $t$.
\end{itemize}

\subsubsection{Regularization:}In the proposed approach, we apply the following regularization strategies during training: (i) We apply dropout to the output of each sub-layer in the attention module prior to residual connections and normalize the outputs. We include an additional dropout layer after adding the positional encoding to the input embeddings, (ii) We also perform attention dropout, similar to \cite{vaswani2017attention}, after computing the self-attention weights.

\subsubsection{Complexity:}Learning long-range dependencies is a key challenge in many sequence modeling tasks. Another notion of complexity is the amount of computation that can be parallelized, measured as the minimum number of sequential operations required. Recurrent models require $O(T)$ sequential operations with a total $O(T\cdot d^2)$ computations in each layer. In comparison, the proposed approach requires a constant $O(1)$ sequential operations (entirely parallelizable) with a total $O(T\cdot r\cdot d)$ computations per layer, where $r$ denotes the size of the mask for self-attention. In all our implementations, $d$ is fixed at $256$ and $r \ll d$, and as a result our approach is significantly faster than RNN training.

\section{MIMIC-III Benchmarks \& Formulation}
In this section, we describe the MIMIC-III benchmark tasks and the application of the \textit{SAnD} framework to these tasks, along with a joint multi-task formulation.

The MIMIC-III database consists of de-identified information about patients admitted to critical care units between 2001 and 2012 \cite{johnson2016mimic}. It encompasses an array of data types such as diagnostic codes, survival rates, and more. Following \cite{harutyunyan2017multitask}, we used the cohort of $33,798$ unique patients with a total of $42,276$ hospital admissions and ICU stays. Using raw data from Physionet, each patient's data has been divided into separate episodes containing both time-series of events, and episode-level outcomes \cite{harutyunyan2017multitask}.  The time-series measurements were then transformed into a $76$-dimensional vector at each time-step. The size of the benchmark dataset for each task is highlighted in Table \ref{table:data-size}. 




\subsubsection{In Hospital Mortality:}Mortality prediction is vital during rapid triage and risk/severity assessment. In Hospital Mortality is defined as the outcome of whether a patient dies during the period of hospital admission or lives to be discharged. This problem is posed as a binary classification one where each data sample spans a $24$-hour time window. True mortality labels were curated by comparing date of death (DOD) with hospital admission and discharge times. The mortality rate within the benchmark cohort is only $13\%$. 

\subsubsection{Decompensation:}Another aspect that affects treatment planning is deterioration of organ functionality during hospitalization. Physiologic decompensation is formulated as a problem of predicting if a patient would die within the next $24$ hours by continuously monitoring the patient within fixed time-windows. Therefore, the benchmark dataset for this task requires prediction at each time-step. True decompensation labels were curated based on occurrence of patient's DOD within the next $24$ hours, and only about $4.2\%$ of samples are positive in the benchmark.

\subsubsection{Length of Stay:} Forecasting length of a patient's stay is important in healthcare management. Such an estimation is carried out by analyzing events occurring within a fixed time-window, once every hour from the time of admission. As part of the benchmark, hourly remaining length of stay values are provided for every patient. These true range of values were then transformed into ten buckets to repose this into a classification task, namely: a bucket for less than a day, seven one day long buckets for each day of the 1st week, and two outlier buckets-one for stays more than a week but less than two weeks, and one for stays greater than two weeks \cite{harutyunyan2017multitask}. 


\subsubsection{Phenotyping:}Given information about a patient's ICU stay, one can retrospectively predict the likely disease conditions. This process is referred to as acute care phenotyping. The benchmark dataset deals with $25$ disease conditions of which $12$ are critical such as respiratory/renal failure, $8$ conditions are chronic such as diabetes, atherosclerosis, and $5$ are 'mixed' conditions such as liver infections. Typically, a patient is diagnosed with multiple conditions and hence this can be posed as a multi-label classification problem. 
\begin{table}[t]
	\centering
	\renewcommand{\arraystretch}{1.3}
	\caption{Task-specific sample sizes of MIMIC-III dataset.}
	\label{table:data-size}
	\begin{tabular}{l|c|c|c}
		\hline
		\multicolumn{1}{c|}{\cellcolor{gray!15}\textbf{Benchmark}} & \cellcolor{gray!15}\textbf{Train} & \cellcolor{gray!15}\textbf{Validation} & \cellcolor{gray!15}\textbf{Test} \\ \hline
		Mortality                                & 14,659                  & 3,244                        & 3,236                  \\ 
		Decompensation                           & 2,396,001                  & 512,413                        & 523,208                  \\ 
		Length of Stay                           & 2,392,950                  & 532,484                        & 525,912                  \\ 
		Phenotyping                              & 29,152                  & 6,469                        & 6,281                  \\ \hline
	\end{tabular}
\vspace{-10pt}
\end{table}

\subsection{Applying \textit{SAnD} to MIMIC-III Tasks}
In order to solve the afore-mentioned benchmark tasks with \textit{SAnD}, we need to make a few key parameter choices for effective modeling. These include: size of the self-attention mask ($r$), dense interpolation factor ($M$) and the number of attention blocks ($N$). While attention models are computationally more efficient than RNNs, their memory requirements can be quite high when $N$ is significantly large. However, in practice, we are able to produce state-of-the-art results with small values of $N$. As described in the previous section, the total number of computations directly relies on the size of the mask, $r$ and interestingly our experiments show that smaller mask sizes are sufficient to capture all required dependencies in 3 out of 4 tasks, except phenotyping, which needed modeling of much longer-range dependencies. The dependency of performance on the dense interpolation factor, $M$ is more challenging to understand, since it relies directly on the amount of variability in the measurements across the sequence. The other hyperparameters of network such as the learning rate, batch size and embedding sizes were determined using the validation data. Note, in all cases, we used the Adam optimizer \cite{kingma2014adam} with parameters $\beta_1=0.9, \beta_2=0.98$ and $\epsilon=10^{-8}$. The training was particularly challenging for the decompensation and length of stay tasks because of the large training sizes. Consequently, training was done by dividing the data into \textit{chunks} of $20000$ samples and convergence was observed with just 20-30 randomly chosen chunks. Furthermore, due to the imbalance in the label distribution, using a larger batch size ($256$) helped in some of the cases. 

\begin{figure*}[t] 
	\label{fig:ph}
	\subfigure[\textit{Phenotyping} - Convergence Behavior]{
		\includegraphics[width=.32\linewidth,clip=True]{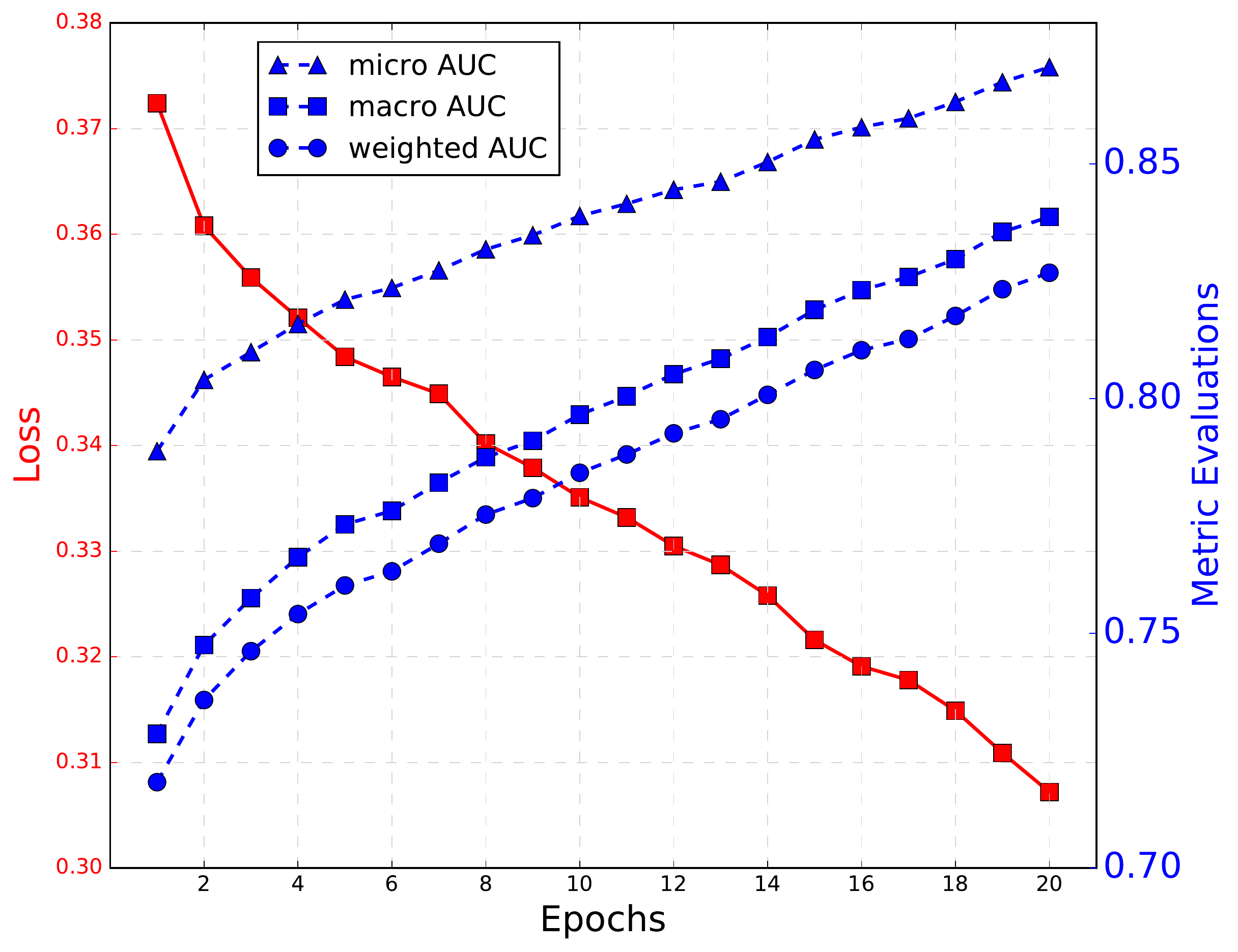}
		\label{fig:ph_loss}}
	\hfill
	\subfigure[\textit{Phenotyping} - Choice of $r$]{
		\includegraphics[width=.31\linewidth,clip=True]{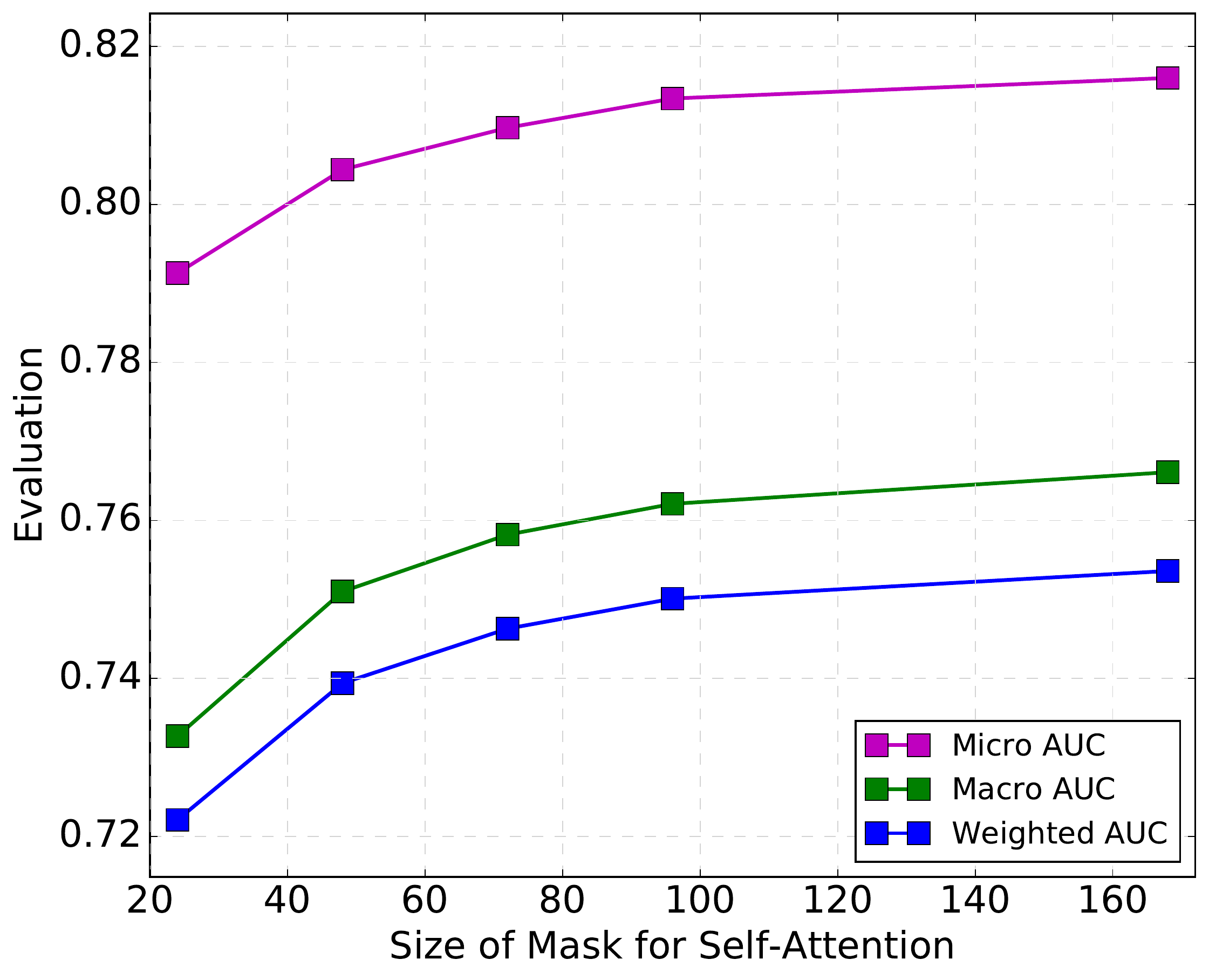}
		\label{fig:ph_r}}
	\hfill
	\subfigure[\textit{Phenotyping} - Choice of $(N, M)$]{
		\includegraphics[width=.32\linewidth,clip=True]{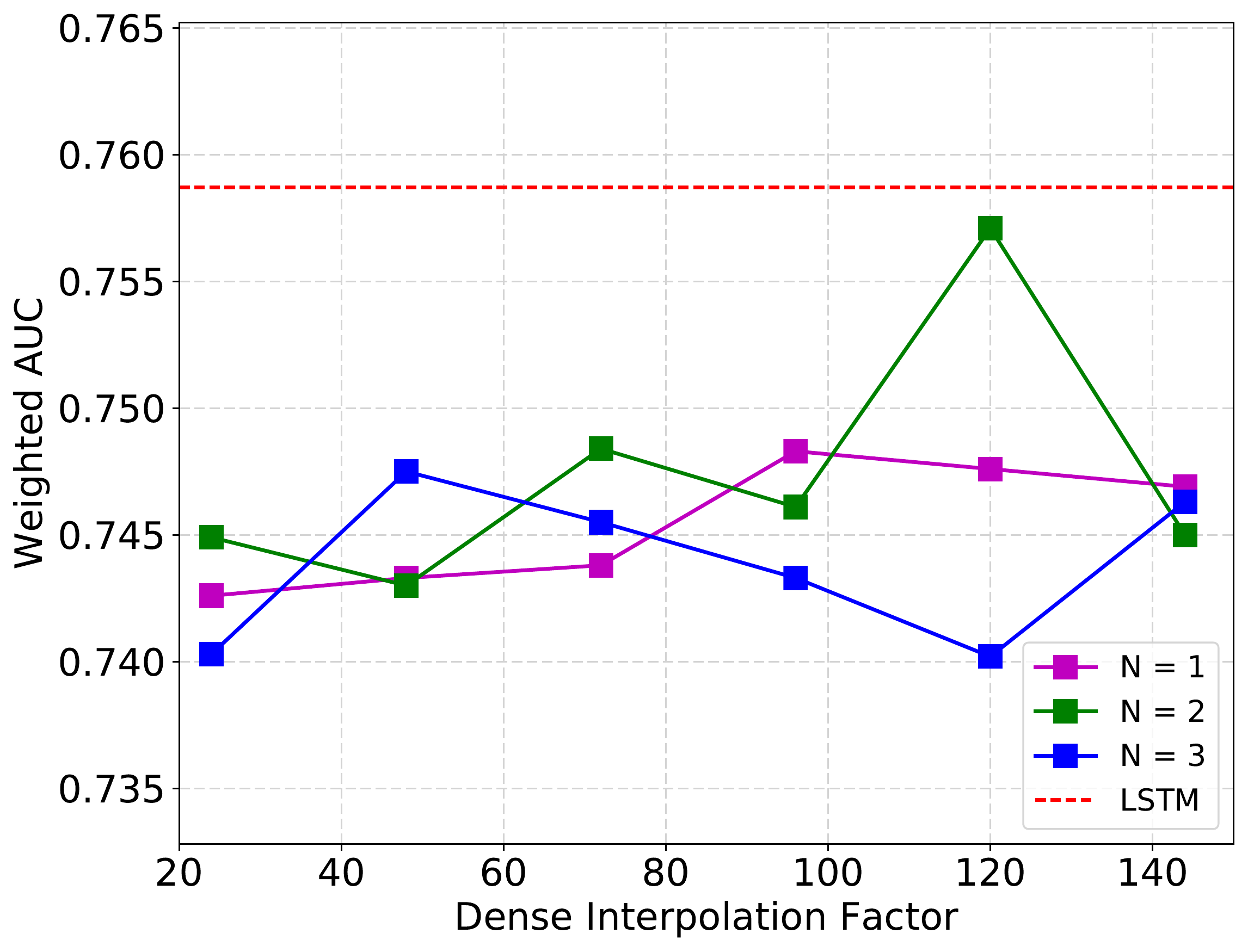}
		\label{fig:ph_nm}}
	\vfill
	\subfigure[\textit{In Hospital Mortality} - Choice of $(N, M)$]{
		\includegraphics[width=.32\linewidth,clip=True]{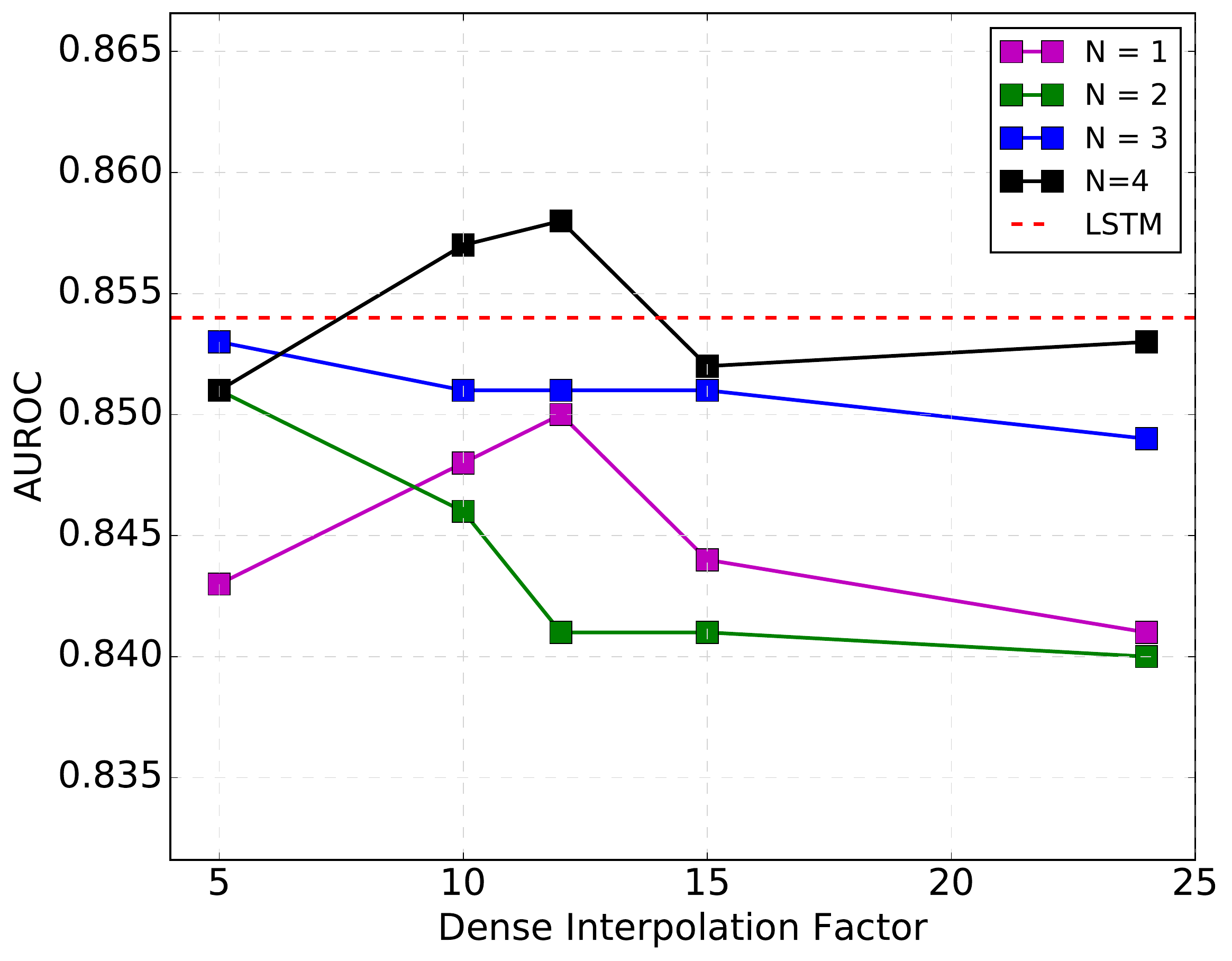}
		\label{fig:ihm_nm}}
	\hfill
	\subfigure[\textit{Decompensation} - Choice of $(N, M)$]{
		\includegraphics[width=.32\linewidth,clip=True]{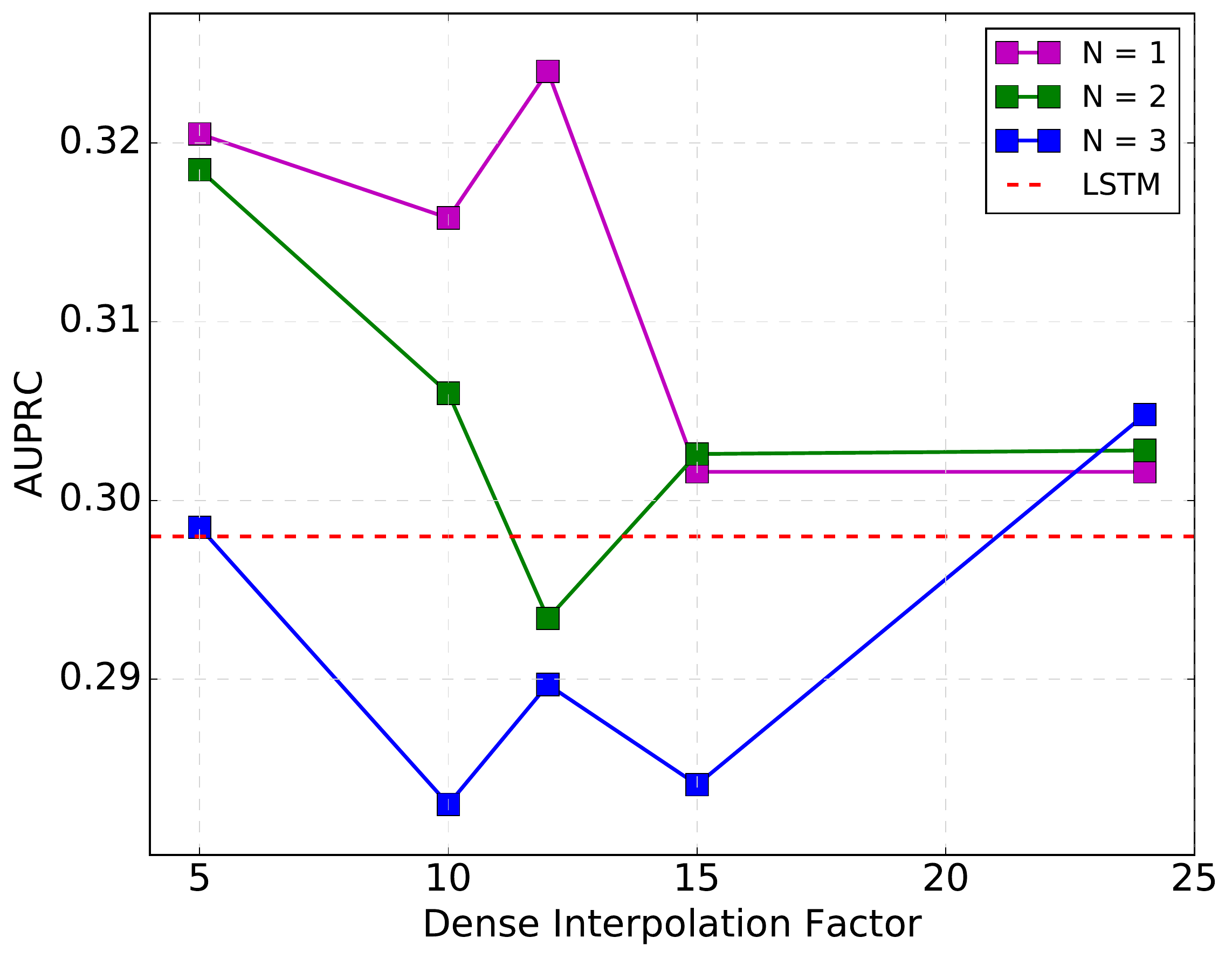}
		\label{fig:dc_nm}}
	\hfill
	\subfigure[\textit{Length of Stay} - Choice of $(N, M)$]{
		\includegraphics[width=.32\linewidth,clip=True]{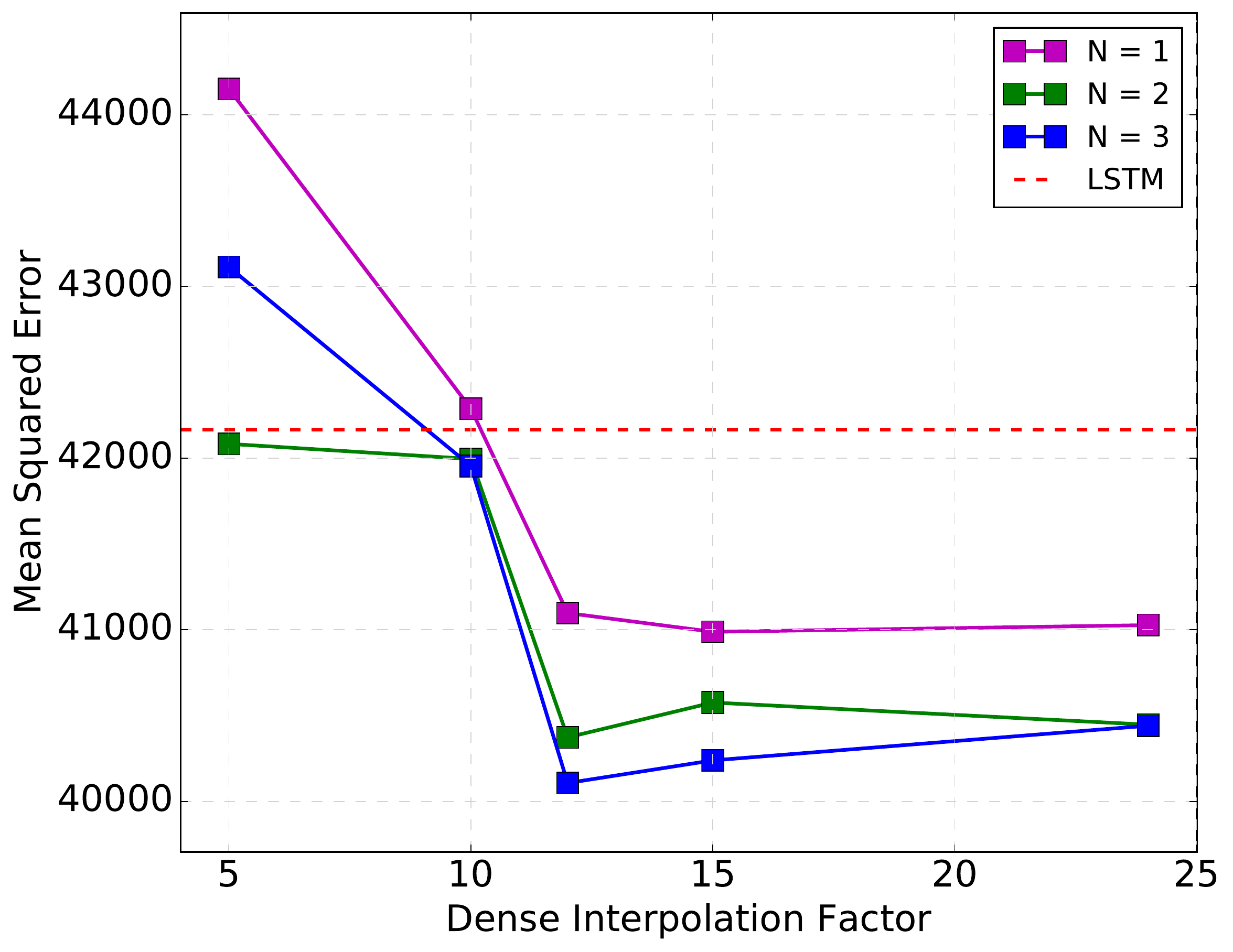}
		\label{fig:los_nm}}
	
	\vspace{-10pt}
	\caption{Applying \textit{SAnD} to MIMIC-III benchmark tasks - We illustrate the training behavior and impact of the choice of the attention mask size, number of attention layers and dense interpolation factor on test performance.}
\end{figure*}

\subsubsection{Multi-task Learning:}In several recent results from the deep learning community, it has been observed that joint inferencing with multiple related tasks can lead to superior performance in each of the individual tasks, while drastically improving the training behavior. Hence, similar to the approach in \cite{harutyunyan2017multitask}, we implemented a multi-task version of our approach, \textit{SAnD-Multi}, that uses a loss function that jointly evaluates the performance of all tasks, which can be expressed as follows:
\begin{equation}
\ell_{mt} = \lambda_p \ell_{ph} + \lambda_i \ell_{ihm} + \lambda_d \ell_{dc} + \lambda_l \ell_{los},
\label{eqn:mt-loss}
\end{equation}where $\ell_{ph}, \ell_{ihm}, \ell_{dc}, \ell_{los}$ correspond to the losses for the four tasks. The input embedding and attention modules are shared across the tasks, while the final representations and the prediction layers are unique to each task. Our approach allows the use of different mask sizes and interpolation factors for each task, but requires the use of the same $N$. 

\section{Performance Evaluation}
In this section we evaluate the proposed \textit{SAnD} framework on the benchmark tasks and present comparisons to the state-of-the-art RNNs based on LSTM \cite{harutyunyan2017multitask}, and baseline logistic regression (LR) with hand-engineered features. To this end, we discuss the evaluation metrics and the choice of algorithm parameters. In particular, we analyze the impact of the choice of number of attention layers $N$, the dense interpolation factor $M$, and the mask size of the self-attention mechanism $r$ on the test performance. Finally, we report the performance of the mutli-task variants of both RNN and proposed approaches on all tasks.

\begin{table*}[t]
	\renewcommand{\arraystretch}{1.3}
	\renewcommand{\tabcolsep}{13.4pt}
	\centering
	\caption{Performance Comparison for the MIMIC-III benchmark tasks, using both single-task and multi-task strategies.}
	\label{table:perf}
	\begin{tabular}{l|c|c|c|c|c}
		\hline
		\multicolumn{1}{c|}{\cellcolor{gray!20}}        & \multicolumn{5}{c}{\cellcolor{gray!20} \textbf{Method}} \\ \hhline{*{1}{>{\arrayrulecolor{gray!20}}-}>{\arrayrulecolor{black}}*{5}{-}}
		
		\multicolumn{1}{c|}{\multirow{-2}{*}{\cellcolor{gray!20} \textbf{Metrics}}} & {\cellcolor{betterblue!40} \textbf{LR}} & {\cellcolor{betterblue!40} \textbf{LSTM}}      &  {\cellcolor{betterblue!40} \textbf{\textit{SAnD}}} & {\cellcolor{betterorange!40} \textbf{LSTM-Multi}}         & {\cellcolor{betterorange!40} \textbf{\textit{SAnD}-Multi}}             \\ 
		\hline 
		\hline
		
		\multicolumn{6}{l}{\cellcolor{blue!10}\textbf{Task 1: Phenotyping}} \\
		
		\hline
		\small
		Micro AUC  & 0.801& \textbf{0.821}        & 0.816   & 0.817 &  0.819   \\
		Macro AUC  & 0.741 & \textbf{0.77}      & 0.766    & 0.766 &  \textbf{0.771}   \\ 
		Weighted AUC& 0.732 & 0.757      & 0.754  & 0.753 & \textbf{0.759}   \\ \hline
		
		\multicolumn{6}{l}{\cellcolor{blue!10}\textbf{Task 2: In Hospital Mortality}} \\
		\hline
		AUROC& 0.845& 0.854   & 0.857 & \textbf{0.863} &  0.859  \\ 
		AUPRC& 0.472& 0.516  & 0.518 & 0.517 & \textbf{0.519} \\ 
		min(Se, P+)& 0.469& 0.491  & 0.5 & 0.499 &  \textbf{0.504} \\ 
		\hline 
		
		\multicolumn{6}{l}{\cellcolor{blue!10}\textbf{Task 3: Decompensation}}\\
		
		\hline
		AUROC& 0.87& 0.895  & 0.895   & 0.900 & \textbf{0.908} \\ 
		AUPRC& 0.2132& 0.298   & 0.316  & 0.319 & \textbf{0.327}  \\ 
		min(Se, P+)& 0.269& 0.344  & 0.354  & 0.348 & \textbf{0.358}  \\
		\hline 
		
		\multicolumn{6}{l}{\cellcolor{blue!10}\textbf{Task 4: Length of Stay}} \\
		\hline
		Kappa& 0.402& 0.427  &\textbf{0.429}  & 0.426 	&  \textbf{0.429} \\ 
		MSE  & 63385 & 42165   & 40373  & 42131 & \textbf{39918} \\ 
		MAPE & 573.5 & 235.9   &167.3  &  188.5 & \textbf{157.8} \\
		\hline

	\end{tabular}
\vspace{-10pt}
\end{table*}

\subsection{Single-Task Case}

\subsubsection{Phenotyping:}This multi-label classification problem involves retrospectively predicting acute disease conditions. Following \cite{lipton2015learning} and \cite{harutyunyan2017multitask}, we use the following metrics to evaluate the different approaches on this task: (i) macro-averaged Area Under the ROC Curve (AUROC), which averages per-label AUROC, (ii) micro-averaged AUROC, which computes single AUROC score for all classes together, (iii) weighted AUROC, which takes disease prevalence into account. The learning rate was set to $0.0005$, batch size was fixed at $128$ and a residue dropout probability of $0.4$ was used. First, we observe that the proposed attention model based architecture demonstrates good convergence characteristics as shown in Figure \ref{fig:ph_loss}. Given the uneven distribution of the class labels, it tends to overfit to the training data. However, with both attention and residue dropout regularizations, it generalizes well to the validation and test sets. Since, the complexity of the proposed approach relies directly on the attention mask size ($r$), we studied the impact of $r$ on test performance. As shown in Figure \ref{fig:ph_r}, this task requires long-term dependencies in order to make accurate predictions. Though all performance metrics improve upon the increase of $r$, there is no significant improvement beyond $r = 96$ which is still lower than the feature dimensionality $256$. As shown in Figure \ref{fig:ph_nm}, using a grid search on the parameters $N$ (number of attention layers) and $M$ (dense interpolation factor), we identified the optimal values. As described earlier, lowering the value of $N$ reduces the memory requirements of \textit{SAnD}. In this task, we observe that the values $N = 2$ and $M = 120$ produced the best performance, and as shown in Table \ref{table:perf}, it is highly competitive to the state-of-the-art results.

\subsubsection{In Hospital Mortality:}In this binary classification task, we used the following metrics for evaluation: (i) Area under Receiver Operator Curve (AUROC), (ii) Area under Precision-Recall Curve (AUPRC), and (iii) minimum of precision and sensitivity (Min(Se,P+)). In this case, we set the batch size to $256$, residue dropout to $0.3$ and the learning rate at $0.0005$. Since the prediction is carried out using measurements from the last $24$ hours, we did not apply any additional masking in the attention module, except for ensuring causality. From Figure \ref{fig:ihm_nm}, we observe that the best performance was obtained at $N = 4$ and $M = 12$. In addition, even for the optimal $N$ the performance drops with further increase in $M$, indicating signs of overfitting. From Table \ref{table:perf}, it is apparent that \textit{SAnD} outperforms both the baseline methods.

\subsubsection{Decompensation:}Evaluation metrics for this task are the same as the previous case of binary classification. Though we are interested in making predictions at every time step of the sequence, we obtained highly effective models with $r = 24$ and as a result our architecture is significantly more efficient for training on this large-scale data when compared to an LSTM model. Our best results were obtained from training merely on about $25$ chunks (batch size = $128$, learning rate = $0.001$) , when $N = 1$ and $M = 10$ (see Figure \ref{fig:dc_nm}), indicating that increasing the capacity of the model easily leads to overfitting. This can be attributed to the heavy bias in the training set towards the negative class. Results for this task (Table \ref{table:perf}) are significantly better than the state-of-the-art, thus evidencing the effectiveness of \textit{SAnD}.

\subsubsection{Length of Stay:}Since this problem is solved as a multi-class classification task, we measure the inter-agreement between true and predicted labels using the Cohen's linear weighted kappa metric. Further, we assign the mean length of stay from each bin to the samples assigned to that class, and use conventional metrics such as mean squared error (MSE) and mean absolute percentage error (MAPE). The grid search on the parameters revealed that the best results were obtained at $N = 3$ and $M = 12$, with no further improvements with larger $N$ (Figure \ref{fig:los_nm}). Similar to the decompensation case, superior results were obtained using $r = 24$ when compared with the LSTM performance, in terms of all the evaluation metrics.

\subsection{Multi-Task Case}
We finally evaluate the performance of \textit{SAnD}-Multi by jointly inferring the model parameters with the multi-task loss function in Eq (\ref{eqn:mt-loss}). We used the weights $\lambda_p = 0.8, \lambda_i = 0.5, \lambda_d = 1.1, \lambda_l = 0.8$. Interestingly, in the multi-task case, the best results for phenotyping were obtained with a much lower mask size ($72$), thereby making the training more efficient. The set of hyperparameters were set at batch size = $128$, learning rate = $0.0001$, $N = 2$, $M = 36$ for phenotyping and $M = 12$ for the other three cases. As shown in Table \ref{table:perf}, this approach produces the best performance in almost all cases, with respect to all the evaluation metrics.

\section{Conclusions}
In this paper, we proposed a novel approach to model clinical time-series data, which is solely based on masked self-attention, thus dispensing recurrence completely. Our self-attention module captures dependencies restricted within a neighborhood in the sequence and is designed using multi-head attention. Further, temporal order is incorporated into the sequence representation using both positional encoding and dense interpolation embedding techniques. The training process is efficient and the representations are highly effective for a wide-range of clinical diagnosis tasks. This is evidenced by the superior performance on the challenging MIMIC-III benchmark datasets. To the best of our knowledge, this is the first work that emphasizes the importance of attention in clinical modeling and can potentially create new avenues for pushing the boundaries of healthcare analytics.


\section{Acknowledgments}
This work was performed under the auspices of the U.S. Dept. of Energy by Lawrence Livermore National Laboratory under Contract DE-AC52-07NA27344. LLNL-CONF-738533.

\bibliographystyle{aaai}
\bibliography{refs}

\end{document}